# Safe AI – How is this possible?[1]

Harald Rueß

fortiss
Research Institute of the Free State of Bavaria for Software-Intensive Systems

ruess@fortiss.org

Simon Burton

Fraunhofer
Institute for Cognitive Systems

simon.burton@iks.fraunhofer.de

Munich, 4th February 2022

*“As we know,*
*there are known knowns.*
*There are things we know*
*we know.*
*We also know*
*there are known unknowns.*
*That is to say*
*we know there are*
*some things we do not know.*
*But there are also unknown unknowns,*
*the ones we don’t know,*
*we don’t know.”*

Donald Rumsfeld, Feb 2002, US DoD news briefing

[1] This work is funded by the Bavarian Ministry for Economic Affairs, Regional Development and Energy as part of the fortiss AI Center and a project to support the thematic development of the Fraunhofer Institute for Cognitive Systems. We are also grateful to Carmen Cârlan and Henrik Putzer for their thorough remarks and suggestions for improvement; in particular, Figure 2 is due to Carmen.

## Table of Content

# 1. Introduction

A new generation of *cyber-physical systems* (CPS) with cognitive capabilities is being developed for real-world control applications. Examples are self-driving vehicles, flexible production plants, automated surgery robots, smart grids, and cognitive networks. These systems are AI-based in that they leverage techniques from the field of Artificial intelligence (AI) to flexibly cope with imprecision, inconsistency, incompleteness, to have an inherent ability to learn from experience, and to adapt according to changing and even unforeseen situations. This extra flexibility of AI, however, makes it harder to predict their behavior, and the difficulty is to construct AI-based systems without incurring the frailties of "AI-like" behavior [1].

In addition, cyber-physical AI systems usually are safety-critical in that they may be causing real harm in (and to) the real world. As a consequence, the central *safe AI* objective is to handle or even overcome the dichotomy between safe and largely unpredictable behavior of complex AI systems.

Consider, for example, an automated emergency braking system for a car that continually senses the operational context based on machine learning (ML), assesses the current situation via an AI decision module based on models of the operational context (and itself), and initiates a maneuver for emergency braking by overriding the human driver, when necessary. The intent of this emergency maneuver is, of course, to prevent accidents in time-critical situations which the human operator may not be able to control anymore. The emergency braking maneuver itself is also safety-related, as wrongful execution might cause severe harm.

The *safe AI* challenge is not exactly new [2] and may well be traced back to Turing himself in the early 1950s. Still, it has recently become all-important due to the euphoric mood on AI, as the acceptance and the success of AI techniques for real-world applications hinges on a meaningful, dependable, and safe control. On-going discussions about the responsible deployment of AI in the real world range from human-centered social norms and values[2] to their robust and safe realization [3] [4].

In this thought outline, however, we restrict ourselves to the technical design and engineering principles for safe AI systems as a necessary step for the responsible deployment of mission- and safety-critical AI systems into our very societal fabric. Moreover, even though we are concentrating in this thought online on safety aspects only, we believe that there is also a fruitful intersection of the suggested approach with related dependability attributes of AI systems such as security, privacy, inverse privacy, fairness, and transparency.

[2] https://www.ai4europe.eu/ethics

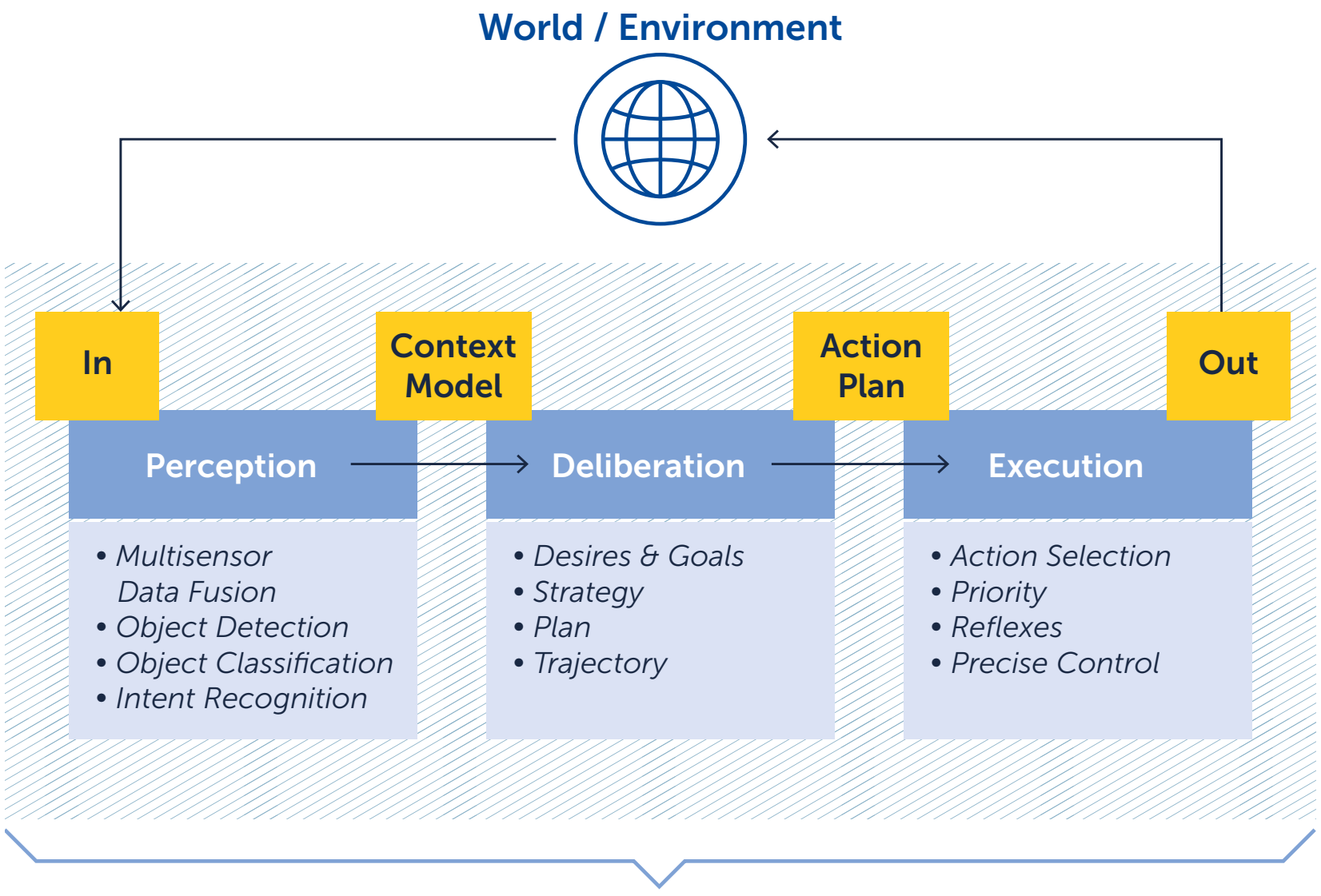

**Figure 1.** Sense-Plan-Act Loop of a Cognitive System.

The starting point of our considerations are cognitive systems, which are software-intensive technical systems that imitate cognitive capabilities such as perception, model-building, and reasoning. More specifically, the basic *sense-plan-ac*t control loop of a cognitive AI system[3] in Figure 1 is based on monitored observations of the operational environment (including the controlled plant), perception, and interacting commands from human operators. Functionally automated driving systems, such as the emergency braking example, may easily be viewed as instances of this *sense-plan-act* loop, where the ego car is the plant to be controlled.

The *cognitive system* in *Figure 1* is conceptually a function taking sensing inputs and generating corresponding output actions, which is usually also based on internal state. Whereas such a loop may be used as the conceptual specification of a reactive CPS [5], it is also the central technical concept of the field of AI, which is concerned principally with designing the internals of stream-transforming controls for mapping from a stream of raw perceptual data to a stream of actions.

Behavior generation for the sense-plan-act loop is decomposed into successive stages for situational awareness, followed by deliberate, goal-oriented planning, and by execution of selected actions in the real world. Sensing functionalities, in particular, are currently often realized through data-driven machine learning methods such as artificial neural networks (ANNs). Behavioral planning capabilities, on the other hand, are usually realized by more traditional software-based control methods, but also through probabilistic and reinforcement-based synthesis of control strategies. Such a conceptual separation into sensing and deliberate planning is supported, among others, by the *global workspace theory*, which categorizes cognitive capabilities into fast and slow modes of operation: System 1 operates rapidly, intuitively, and effortlessly, whereas System 2 requires concentration, motivation, and the application of learned rules, and it allows us to grasp the right ones.[4] In other words, System 1 means snap judgements which seduce us with the wrong answers, and System 2 means thinking twice [6].

[3] For our purposes, we use the terms “AI system”, “cognitive AI system”, and “cognitive system” largely interchangeably.

The context model of our running example, automated emergency braking, might consist of the positions, bounding boxes, and motion vectors of surrounding environment objects such as cars or cyclists. The *sensing* stage constructs and updates faithful models,[5] based on perceived inputs and other knowledge sources, of both the exogenous operating environment and the endogenous *self*. One can easily imagine scenarios in which failure of detection, misclassification, or imprecision in models is the cause of an accident. The main challenge therefore is to provide a convincing argument that an AI system is *sufficiently safe* as determined through applicable risk and safety analysis. As usual, this notion of *sufficiently safe* heavily depends on the specific societal context and, correspondingly, acceptable risks.

For our purposes, automated emergency braking, say, EB is intuitively said to be *safe* if its activation prevents, at least up to some tolerable quantity, accidents in prescribed situations. Assuming we can identify a corresponding subset S of "known" safe states of the operating context, the *safety envelope*, then the safety challenge for EB reduces to verifying the safety invariant $EB(S) \subseteq S$. In this way, emergency braking EB, when initiated in a potentially unsafe and uncontrollable (for the driver) state in S, produces safe control actions, in that the ego car is always maneuvered towards a safe and controllable state, possibly a fail-safe state, and as the basis for a possible handover to the driver. As with most CPS, ensuring safe control involves a rather complex interaction of uncertain sensing, discrete/probabilistic computation, physical motion, and real-time combination with other systems (including humans). We are arguing that traditional safety engineering techniques for embedded systems and CPS is, for the multitude of heterogeneous sources of uncertainty, not applicable to learning-enabled cognitive systems which are acting increasingly autonomous in open environments.

We identify central specification, uncertainty, assurance, design, analysis, and maintenance challenges for realizing such a rigorous design of safe AI; all based on the notion of managing uncertainty to acceptable levels.[6] An overview on these challenges is provided in Table 1 - without any claim of completeness.

Also notice that, due to the infancy of safe AI engineering, at times this exposition may seem to be rather sketchy and speculative, and clearly, many of our claims and hypotheses need further substantiation or disproval. In this sense, this thought outline is supposed to be both provocative and thought-inspiring. It is also intended to be a living document, which needs to be updated and concretized as we gain more experience and increase our theoretical understanding on the rigorous design for safe AI – as the basis for the responsible and safe deployment of AI in our economic and societal fabric.

[4] Notice, however, that there are exceptions to this suggested separation-of-means, such as Nvidia's end-to-end-control for an experimental self-driving systems [119].

[5] So-called digital twins.

[6] In analogy to the "as low as reasonably possibly" (ALARP) risk-based criterion we might call this the "as certain/confident as reasonably possibly" (ACARP) principle.

# 2. Challenges

## Uncertainty and Complexity

The cognitive capabilities of cyber-physical systems are enabled by advances in AI, in particular ML as well as the large-scale availability of training and validation data through an increasing number of sensing channels and connectivity. As motivated above, the deployment of such systems is leading to significant challenges in safety assurance including such existential statements such as whether AI systems can ever be considered safe enough. We now look at some of the legitimate reasons for these doubts before focusing on the AI-specific topics, exploring them in more detail in later sections.

Previously, safety-critical electric/electronic (E/E) systems were assured by considering the impact of malfunctions caused predominantly by either random hardware failures or system design faults, including, but not exclusive, to software bugs. This allowed for a model-based approach to understanding the failure modes of individual components and how faults in individual components propagate through the system leading to hazardous actions. However, the introduction of safety-critical cognitive systems requires a broader consideration of safety and potential causes of hazards. Many of these challenges can be related to the increasing *complexity* and *uncertainty* within both the system and its environment.

**Uncertainty.** A particular challenge is that there are a multitude of sources of entangled uncertainty in an AI system. The inductive capability of ML for extracting models from data is inseparably connected with uncertainty, but there is also uncertainty about the operating context, there is uncertainty about the models of the operating context and the "self", there is behavioral uncertainty due to the approximate nature of heuristic learning algorithms, there is uncertainty due to probabilistic and non-deterministic components, there is uncertainty about safety hazards,[7] there is uncertainty about safety envelopes in uncertain operating contexts, there is uncertainty on a meaningful fallback to a responsible human operator, and, finally, there is also uncertainty in self-learning systems about their emergent behavior in time. Possibly the only thing that is certain about an AI system is that it is uncertain and largely unpredictable.

Let's investigate, for example, sources of uncertainty of ML components such as artificial neural networks (ANNs) in more detail. The input-output behavior of ANNs heavily relies on the selection of "complete" and "correct" (with respect to the ground truth) sets of training and support data for faithfully specifying relevant operating contexts (input) and their intended internal representation (output). Another source of uncertainty for these ML algorithms is due to the use of stochastic search heuristics, which may lead to incorrect recall even for inputs from the training data, and the largely unpredictable capability of generalizing from given data points. Uncertainty on the faithfulness of the training data representing operating contexts and uncertainty on the correctness and generalizability of training also combine in a, well, uncertain manner. The consequences of these accumulated uncertainties are profound. Particularly, ANNs are usually not robust on unseen inputs, as there is also quite some uncertainty on the behavior on even small input changes.[8]

[7] For instance, dynamic hazards such as the sudden occurrence of objects on the road which may lead to catastrophic failure.

Adequate approaches are needed for measuring (un)certainty of the input-output behavior of an ANN with respect to the real world. For example, how certain are we that a given ANN correctly classifies certain classes of homeomorphic images of a “cat”? How certain should an ego vehicle be that there are no surprises, such as undetected or misclassified vehicles, before initiating an emergency brake? Based on these certainty measures, internal models of the operating context should be equipped with confidence levels or, more generally, confidence intervals or distributions. The emergency braking assistant, for example, may, as the basis for selecting appropriate action, assign confidence levels for the position and mobility vectors of all relevant objects. Possibly together with a confidence level that objects have been correctly identified and classified and that there are no “ghost” objects in the context model.

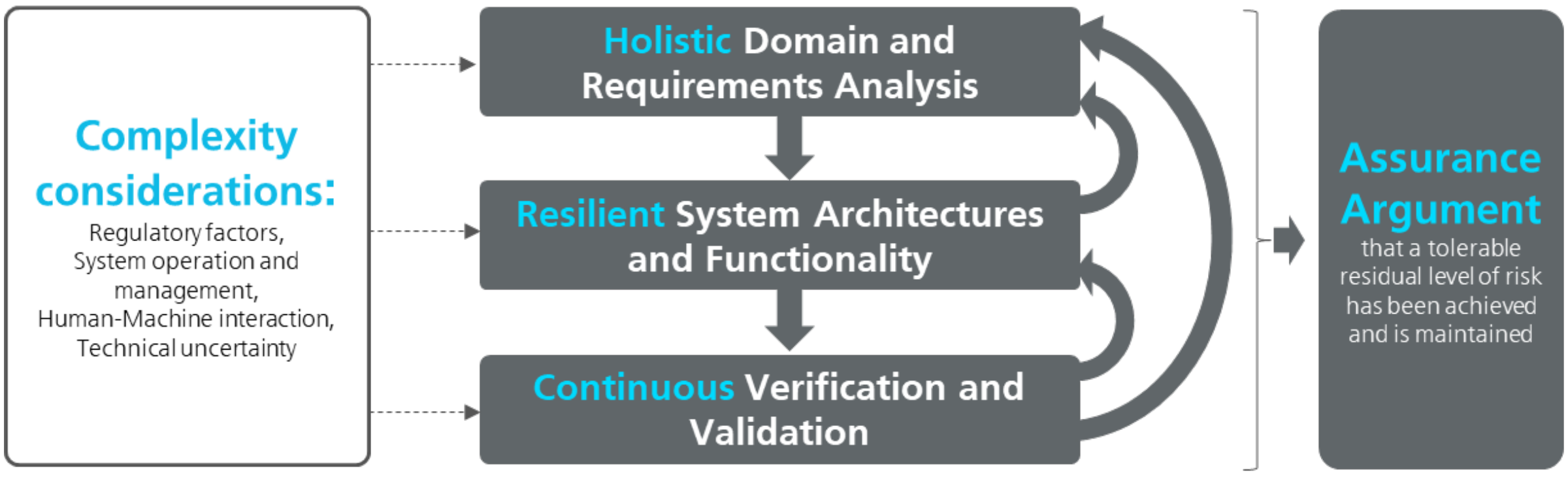

*Figure 2: Complexity-aware systems safety engineering.*

**Complexity**. We refer to complexity in terms of *systems theory,* where a system is defined as complex if the interaction between parts of the system lead to behavior that could not be predicted by considering the individual parts and their interactions alone. Complexity can manifest itself within different levels of the system for example:

- **Increasing complexity within the E/E architecture.** This is caused not only by the increasing number of technical components within a system, but also the heterogeneity and technical implementation of these components, the use of components and software of unknown pedigree and changes in the system after release due to software updates or the integration of additional services (e.g. via cloud connectivity). One impact of system complexity is non-linearity, mode transitions and tipping points where the system may respond in unpredictable ways depending on its current state or context.

[8] For instance, “one pixel attacks” for fooling deep neural networks [22].

- **Complex behavioral interactions between systems, self-organization and *ad hoc* systems-of-systems.** Interactions between the system and its environment may be difficult to predict, especially when human agents are involved in the interactions. Consider the range of behaviors that must be considered by a self-driving vehicle navigating heavy traffic consisting of human-driven vehicles as well as automated vehicles from other manufacturers acting according to unharmonized norms of behavior. Such interactions *may* lead to *ad hoc* systems-of-systems forming, over which individual manufacturers have no or little control and thus call into question whether the system scope under consideration for safety is appropriate and what an appropriate scope of analysis should be.

Increasing complexity makes it more difficult to determine the (potential) causes of failures in the system and effective risk control measures. The impact of complexity in the systems and their environment on our ability to deliver convincing arguments for safety has been discussed in more detail within the scope of automated driving [7]. The concept of uncertainty is closely related to the topic of increasing complexity. Again, uncertainty can manifest itself in several ways that make the safety assurance of safety-critical cognitive systems more challenging.

1. **Scope and unpredictability of the operational domain.** Many highly automated cyber-physical systems can be said to operate within an open context. That is, an environment which cannot be fully specified in a way that desirable system behavior can be defined for each possible set of conditions. Such environments are typified by the presence of edge cases, or “black swans”, corresponding to previously unknown or even unknowable conditions. Furthermore, the operational domain can shift over time leading to new sets of conditions which were not considered during design. This inevitably leads to insufficiencies in the resulting specification of the system under development, which are referred to as “ontological” uncertainties [8].

2. **Inaccuracies and noise in sensors and signal processing.** This complex, unpredictable environment is measured using a combination of inevitably imperfect sensors providing a noisy, incomplete view of the environment. In additional to general inaccuracies in the measurements, such sensors themselves can be “fooled” by physical properties of the environment, such as lens flare distorting a video image or manhole covers leading to spurious radar reflections.

3. **Uncertainties in the perception and decision-making functions.** The complex, incomplete and noisy inputs to the system are often the motivation for using AI and machine learning techniques in the first place. However, as we will explain in further chapters, these algorithms themselves introduce additional uncertainty within the system and rarely deliver precise results. Therefore, in an attempt to solve the problem of uncertainties in the inputs to the system, yet another class of uncertainties are introduced.

The complexity of the system and associated uncertainties lead to *semantic gaps* [9], which are defined as a discrepancy between the intended and specified functionality and can be caused by the complexity and unpredictability of the operational domain, the complexity and unpredictability of the system itself as well as the increasing transfer of decision functions to the system which would otherwise require non-specifiable properties such as human intuition or ethical judgement. These semantic gaps lead to insufficiencies in the definition of appropriate safety acceptance criteria as well as a lack of confidence that statements made in a safety assurance case reflect the actual achieved safety of the system.

The above discourse illustrates the manifold challenges we face when developing safe cognitive systems. It also allows us to better delimit discussions around “Safe AI”. To derive an adequate set of safety assurance methods for such systems we must be clear about which problems we are addressing. These can be roughly separated into the following categories:

1. **Safety challenges caused by the inherent difficulty of the task to be solved.** This includes the systematic complexity of the function to be implemented using AI components based on the complexity and unpredictability of the input domain and the resulting impact on semantic gaps which may restrict our ability to define an adequate specification of the required performance of the AI-based function. These factors are independent of the actual AI or ML techniques used and are better referred to as cognitive systems safety engineering activities. This includes the application of suitable systems safety assurance methods, including the definition of socially and legally tolerable risk acceptance criteria as the development of an overall system design that is resilient to previously unknown or changing properties of the environment A “complexity-aware” systems safety engineering approach is summarized in
2. Figure **2**.

3. **Safety challenges caused by the use of specific AI/ML techniques.** This includes performance limitations and properties of the specific AI/ML-techniques used. For example, statistical modelling and linear regression-based models exhibit different sets of properties related to the explainability and predictability of their results as deep neural networks but may differ greatly regarding their accuracy for certain tasks. For an example on how properties of the specific ML-technique can support a safety assurance case can be found in [10]. The AI-technology specific challenges therefore involve ensuring that the specific performance requirements allocated to the AI-based function within the system context are fulfilled with a level of confidence commensurate to the overall level of system risk.

When discussing "Safe AI" and associated challenges, we should therefore be clear which scope we are referring to. Are we referring to the safety of cognitive systems operating within an open context or whether specific properties of a trained model remain within certain bounds of uncertainty for a given set of inputs? The two topics are closely inter-related. For example, when applying ML techniques such as Deep Neural Networks that deliver a high level of prediction uncertainty or sensitivity to small changes in the inputs, then the cognitive systems engineering task must ensure that tolerances on uncertainties within the trained model must be carefully defined and aligned with other system components.

## Safety Engineering

Traditional safety engineering ultimately is based on fallback mechanisms to a responsible human operator, deterministic behavior of the cyber-physical system as the basis for its testability, and well-defined operating contexts.  In addition, current safety certification regimes require correct and complete specifications prior to operation. These basic assumptions of safety engineering do not hold anymore for AI systems, as:

1. With increased autonomy a fallback mechanism to a human is often not possible anymore. Indeed, the emergency braking system needs to perform without any human intervention as the required reaction times are well below the capabilities of human beings.

2. AI systems make their own knowledge-based judgements and decisions. While added flexibility, resilience, elasticity, and robustness of cognitive AI systems are clearly important, the gains in these dimensions come at the loss of testability due to the admittance of non-determinism.[9] This is a high price to pay, since systematic testing and simulation currently still is the single most used technique for verifying the correct functioning of software-intensive systems.

3. AI systems need to cope with operating environments that cannot be comprehensively monitored and controlled, and in which unpredictable events may occur.  In fact, the main reason of using AI systems is for those situations where the full details of the operating context cannot be known ahead of time. It is therefore difficult to carry out risk estimation for AI systems using conventional techniques.

[9] A NASA study of a software-based control of vehicle acceleration, for example, revealed, among others, potential race conditions in sensor readings due to asynchronous access by a multiplicity of threads. This study concluded that the software was "untestable", making it impossible to rule out the possibility of unsafe control actions [124].

For all these reasons, well-established and successful safety standards for software-intensive systems, including DO 178C in aerospace and ISO 26262 in the automotive industry, cannot readily be applied to AI systems. Indeed, these safety standards pay little attention to autonomy and to the particularly advanced software technologies for system autonomy [11].

There are many on-going industrial initiatives for developing, for example, low-level functionally automated driving[10] and certified AI algorithms in the medical domain.[11] These endeavors, however, are incomplete in that they are based on prescriptive safety standards.[12] But we still need to figure out adequate methodologies and end-to-end verification technology for assuring safe autonomy. And we also need to gain more practical experience on these approaches before prescribing them as "good" or even "best" practice in industrial standards. Current safety engineering standards are also based on the idea that the correct behavior must be completely specified and verified prior to operation. It is therefore unclear if and how these safety standards may apply to learning-enabled systems, which are continuously self-adapting and optimizing their behavior to ever-changing contextual conditions and demands, based on their experience in the field.

If current safety engineering methodology are not directly applicable to AI systems, then we might ask ourselves if we can at least reduce problems of safe and learning-enabled control on a case-by-case basis; for example:

1. Depending on the application context, safety engineers can restrict AI-based functionality with the intent of increasing controllability or decreasing severity/exposure, thereby decreasing associated safety risks.

2. Uncertainty due to open-ended operating contexts and safe control thereof is dramatically reduced in current automotive practice, by collecting all kinds of possible driving scenarios by means of global ecosystems of vehicles – both real and virtualized.[13] This approach basically tries to close-off the set of possible driving scenarios as the basis for constructing the equivalent of a "digital rail". Due to the large number of possible anomalies ("black swans"),[14] however, this approach is rather resource-intensive, and it is unclear how to determine that enough scenarios have been collected to sufficiently cover the space of all driving scenarios.

[10] Safety First guidance for potential methods and considerations with intention of developing safe L3-L4 automated driving functionality including ANN [105].

[11] https://www.fda.gov/medical-devices/software-medical-device-samd/artificial-intelligence-and-machine-learning-aiml-enabled-medical-devices#resources

[12] "If you do … then the system is safe."

[13] For example, https://www.pegasusprojekt.de/en/home

[14] A well-known example is the fatal crash of a car in autonomous mode that resulted from a very rare four-factor combination of a white truck against a brightly lit sky, along with truck height and angle versus the car (https://www.tesla.com/blog/tragicloss), (https://www.ntsb.gov/news/press-releases/Pages/PR20170912.aspx)

3. AI-based functionality is complemented by a safe control channel, thereby effectively combining the intended performance of AI-based systems with the safety of more traditional control.[15] The crucial element in such a safety architecture is a switch between the performant AI-based channel and the safe channel, which is based on run-time monitoring of crucial safety specifications. A pervasive runtime monitor, for example, checks that the proposed action of *EB* yields safe behavior. Such a pervasive runtime monitor together with the switching logic between the two channels is developed with traditional safety engineering methods, thereby effectively removing AI component from a safety-critical path.

4. An engineer may also decide to discontinue initial AI-based proof-of-concepts by reverting to well-understood control techniques altogether; for example, if an end-to-end safety concept for the given AI-based functionality is too costly or not possible, and if a sufficiently performant and safe system may be achieved by more traditional means. The underlying phenomenon of *technical debt* for data-driven systems in real-life engineering has already been described previously [12].

Based on these kinds of engineering design decisions for reducing the safety relevance of AI systems and for increasing its determinacy, it may at times be possible to responsibly use traditional prescriptive safety engineering techniques also for AI. However, this reductionist approach is restricted to a rather small class of functionally automated systems with some added machine learning-based capabilities, which do not adequately support the key concepts of AI systems, namely, autonomy and self-learning. Methods for reducing safe AI problems to the currently prevailing prescriptive safety engineering standards therefore are not future proof,[16] since prescribed and fixed verification and validation process activities, criteria, and metrics does not work well for assuring the safety of AI systems [13].

Overarching Properties [14] have recently been proposed as a product-based alternative to prescriptive safety engineering standards such as DO 178C. Informally, a system is safe for operation if and only if the CIA conditions hold:

1) the system does what it is supposed to do under foreseeable operating conditions (**C**orrectness);
2) what the system is supposed to do is properly captured (**I**ntent); and
3) the system does not cause unacceptable harm (**A**cceptability).

[15] This is sometimes called a "Simplex" architecture.
[16] The applicability and limits of prescriptive safety standards to autonomous AI systems is also discussed in [126].

An assessment of whether a system possesses these properties might be based on an explicit assurance case.[17] Overarching properties therefore are flexible enough to be adapted to developing justified belief of system safety with learning-enabled components. As of now, however, the overarching CIA properties do not seem to have been adopted for safety certification on a larger scale.

All taken together, traditional safety engineering is coming to a turning point moving from deterministic, non-evolving systems operating in well-defined contexts to self-adaptive and self-learning systems which are acting increasingly autonomous and in largely unpredictable operating contexts. But we currently do not have an adequate safety engineering framework for designing this upcoming generation of safety-related AI systems.[18]

In the following we therefore outline a novel approach for safe AI engineering. It is based on *uncertainty quantification* for the multitude of sources of uncertainties of AI systems. The overarching goal of managing uncertainties is to minimize uncertainty in the system behavior, thereby increasing confidence, up to tolerable levels, in the safe behavior of the AI system.

The underlying idea is to generalize the notion of *determinacy* in traditional safety engineering to *uncertainty*. As a special case, if there is no uncertainty on the behavior of some system, then it is fully predictable and deterministic. Deterministic parts of an AI system can (and probably should) therefore still be developed with well-proven design and verification techniques for establishing correctness and possible perfection.

Instead of fallback to responsible (human) operators, uncertainty measures are used by autonomous control strategies for minimizing surprises and for safely exploring largely unknown territory. The emergency braking assistant, for example, might be rather uncertain on the precise location of some relevant car, and it therefore initiates additional perceptive capabilities with the intent of decreasing uncertainty, thereby increasing its confidence, on the location of the respective car to a sufficient level; as the basis for deciding on a safe sequence of emergency braking actions. Moreover, some indirect *cues* [15] cause the system to hypothesize the existence of a relevant car, which, needs to be confirmed by additional actions before initiating emergency breaking. These examples demonstrate that uncertainty is not only a design but also an essential runtime artifact for the situational generation of safe control behavior.

Uncertainties in the proposed engineering framework are explicitly managed through *safety cases*. These are structured arguments, supported by a body of evidence that provides a compelling, comprehensible, and valid case that a system is safe for a given application and operating environment.[19] In contrast to largely *process-based* traditional prescriptive approaches, a safety argument based on safety cases is largely *product-based*. As it involves the presentation of evidence that the actual developed system is safe, as opposed to merely showing that it was developed using normatively prescribed “good” practice. Recent quantitative extensions to safety and assurance cases provide the basis for assigning and combining uncertainties for the central ingredients, such as evidence, arguments, assumptions, and conclusions.

[17] See also Subsection *Assurance-based uncertainty estimation*.
[18] See, for example, [129] for a survey on safety certification of systems with learning-enabled components.
[19] Def Stan 00-56 Issue 3, Safety Management Requirements for Defence Systems, Ministry of Defence, 2007.

Figure 3, for example, illustrates the top-level plan for constructing a safety case for an autonomous GNC,[20] which is built upon a traditional 3-level autonomous architecture. The modular construction of this safety case is based, among others, on evidence from traditional verification of planning components, verification of the correctness or quasi-predictability of neural network components for perceptive tasks, and runtime monitoring for central safety properties. This GNC also includes an FDIR[21] component for detecting and for recovering from unforeseen and potentially hazardous events. The overall goal is to develop an autonomous spacecraft, say, for landing on an asteroid together with a complete safety case. And the safety case is also used in generating safe landing behavior and safely handles unforeseeable events.

[20] **G**uidance, **N**avigation, and **C**ontrol
[21] **F**ailure **D**etection, **I**solation, and **R**ecovery

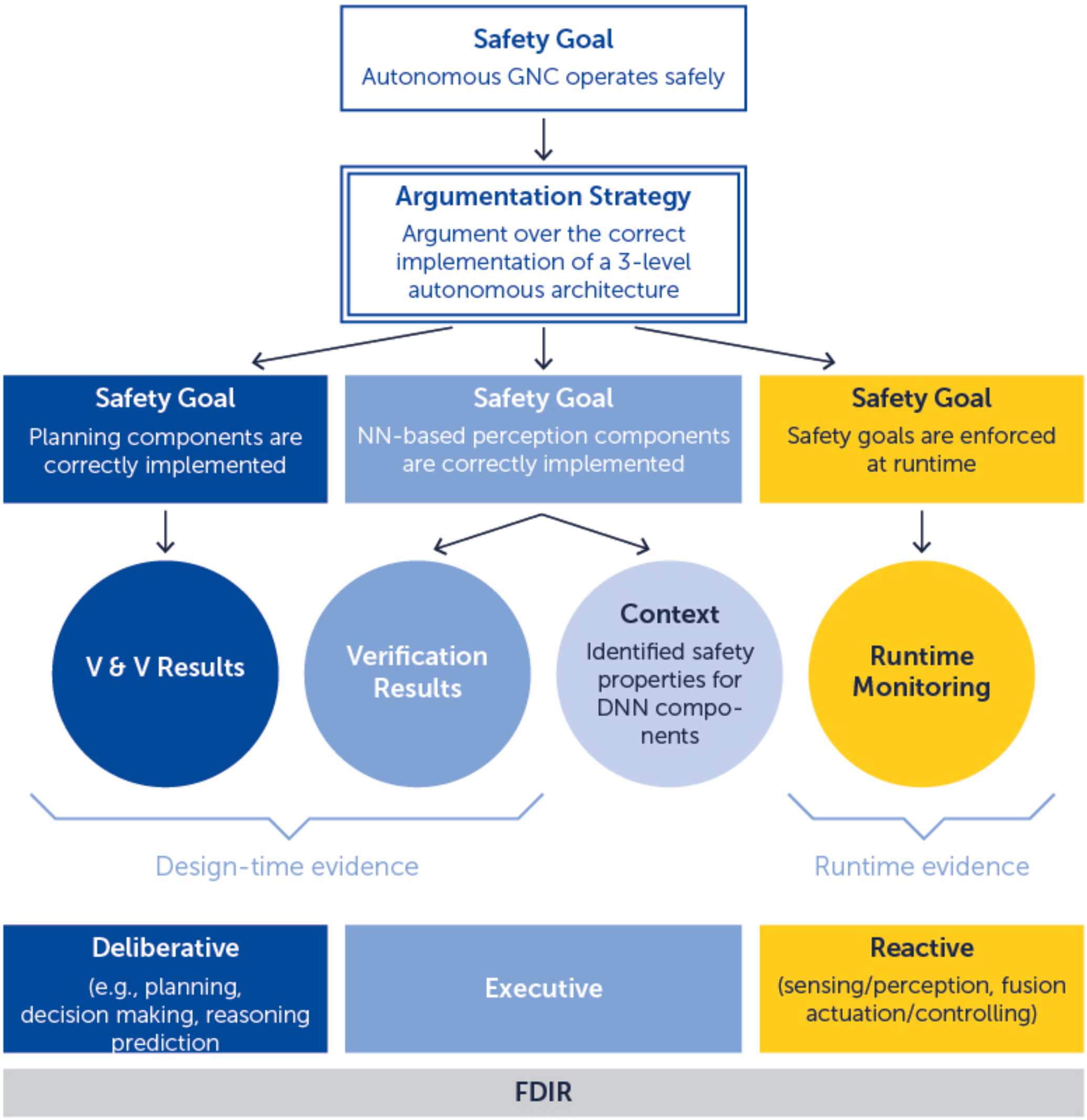

***Figure 3.*** Safety case from a 3-layer architecture of a cognitive system.

## 3. Specification

In a first step we need to express in precise terms when an AI system may be considered safe. This sounds easier than it is, since data-driven AI is particularly successful in application contexts for which it is difficult, if not impossible, to come up with concise specifications (say, translation of natural language). Also operating contexts of AI systems typically are complex, uncertain, and largely unpredictable. Safety hazards are subject to change during operation, and the presence of human operators and their interaction with the sense-plan-act control loop further complicates matters in that even well-intended human interaction may lead to unsafe behavior. In summary, the adaptive, flexible, and context-sensitive behavior of AI systems causes unpredictability and emergent unsafe behavior, which was not necessarily specified, intended, or consciously implemented during system design.

In the following we distinguish between safety specifications of AI systems and derived specifications of learning-enabled components of these systems. And we also discuss ways for systematically deriving requirements for individual components of the system from overall safety requirements.

## System Safety Specification

The overall system's safety specification is often described in terms of safety envelopes. These envelopes may be thought of as under-approximation of the states (or scenarios) of possible operating contexts that are sufficiently safe.[22] In a slightly more general setting, we also quantify uncertainty of environment states.

A common approach for specifying safety envelopes is based on maximizing under-approximations, thereby also maximizing the number of known safe behaviors. In other words, the operating context is partitioned into known safe (the safety envelope), the known unsafe, the unknown safe, and the unknown unsafe, and the goal is to maximize the known safe areas by minimizing the known unsafe areas and discovering as many new unsafe scenarios as possible with a given level of effort [16].[23] And then, there is one more thing, as safety envelopes need to be continually adapted to ever-changing operating contexts, safety hazards, and safety requirements.

The safety objective of the automated emergency braking system, for example, is to maintain a minimum safe distance between the ego car and environment objects. More generally, the Responsibility-Sensitive Safety (RSS) model offers provable safety for vehicle behaviors such as minimum safe following distance [17]. Safety envelopes such as RSS have also been extended to deal with worst-case variability and uncertainty [18]. Safety envelopes usually are highly non-linear and context dependent as is the case for Kamm's friction ellipses. In these cases, machine learning techniques based on minor component analysis are promising for synthesizing safety envelopes from safe behaviors [19]. These techniques open the possibility of self-learning and safe maintenance of safety envelopes through cautious[24] and safe exploration.

We still have very little knowledge on how to systematically construct safety envelopes. The challenge therefore is to construct and maintain safety envelopes, that is known safe states, of the operational context, and to maximize the safety envelope with a given level of effort. There is uncertainty on the safety of certain states. Operational contexts, system behavior, and the notion of acceptable risk are constantly evolving. Therefore, safety envelopes need to be adapted continuously to these ever-changing conditions and requirements. Depending on the degree of autonomy, such AI systems need to self-maintain corresponding safety envelopes, possibly including online risk analysis.

[22] According to a risk and safety analysis of the system under consideration
[23] A newer version is available at https://www.beuth.de/de/norm-entwurf/iso-dis-21448/335355102
[24] Cautious behavior might be realized by means of *minimizing surprises*, which may realized by minimizing *free energy* or maximizing *predictive information*.

## Component Safety Specifications

The behavior of machine learning components is specified by means of data. In supervised learning of ANNs, for example, a set of interpolation points is used for specifying the input-output behavior of the intended function. Inputs may be observations of the operational context, and outputs are corresponding context models.

The central challenge for data-driven requirements engineering is to represent the operational context by means of selected interpolation points as faithfully as possible for a given level of effort. This can be accomplished by sampling input scenarios from the (assumed) distribution of the operational context or by discretizing the operational context according to n features into $2^n$ cells and sampling inputs from these cells. Both approaches are prohibitively expensive for many interesting operational contexts. A series of polynomial-time approximations, for example, have been developed to make feature-based discretization feasible in practical applications [20], and clustering and unsupervised learning techniques are used for identifying a finite number of representative classes of scenarios [21]. It may also be interesting to construct sets of interpolation points that, when used for training a perception ANN, are sufficient for establishing invariance of the AI system with respect to a given safety envelope.

Given a set of, hopefully, representative interpolation points and an initial neural network architecture, an ANN is constructed by heuristically searching, usually based on hill-climbing, for a configuration to minimize the error between actual and specified outputs. This kind of heuristic search might get stuck in local minima, thereby leading to suboptimal solutions, or it might not terminate at all. Moreover, the resulting ANN might be *incorrect* in that its input-output behavior does not coincide with the specifying interpolation points, and the ANN might not be *resilient* in that slight variations of inputs lead to completely different output behavior, since many ANNs have a tendency of “overfitting” without further precautions in training. Consequently, the output of ANNs can often be altered by adding relatively small perturbations to the input vector [22].

These kinds of uncertainties motivate the need for requirements of ANNs beyond data, such as resilience. For a fixed input, say $x$, and a metric *d* on the input space, an ANN is locally $\varepsilon$ -resilient if ANN($x'$) is equal to – alternatively: similar to – the output ANN($x$) for all small perturbations $x'$ *with* $d(x, x') < \varepsilon$; if the ANN is locally $\varepsilon$-resilient for all possible inputs, then it is also globally $\varepsilon$-resilient [23] [24]. The inherent uncertainty on the input-output behavior of ANNs is considerably reduced by establishing strong resilience properties.

Additional desired behavior of perception components usually come with their intended functionality. For example, an object classifier may be expected to correctly classify certain affine or homeomorphic images of training inputs, such as stretching, squeezing, rotational, and translational images. Sequences of context models for modelling traffic flow, for example, are also expected to obey fundamental laws of physics.

The challenge is to identify and maintain desired properties and potential defects of ANN-based perception components, which lead to undesired behavior. Moreover, the contribution of those properties and defects on the overall system safety need to be understood much better. This kind of knowledge should enable the development engineer, for example, to compute precise bounds on the required resilience of the perception ANN for arguing overall system safety.

## Deriving Component Safety Specifications

Given a safety specification, say S, of an AI system as depicted in Figure 1, we derive corresponding safety constraints on the possible behavior of the ANN-based perception component. The control, consisting of the deliberation and the execution units, needs to ensure that the output (the changed environment) is safe, that is, in S. In an engineered system we can compute the weakest precondition, say, $wp(controller)(S)$, which now serves as the post-condition for the perception unit. Assuming that the input x to the perception unit is in S, that is, this state is safe, we get a pre/post-condition specification

$$(\forall\, x\ \in S)\ perception(x) \in wp(controller)(S)$$

of the perception unit which is sufficient to establish safety of the overall AI system loop.[25] Adequate domain abstractions and corresponding abstract interpretation techniques are needed to make this approach feasible.

Indeed, researchers have taken first steps in this direction and identified special cases of pre/post-condition pairs for neural networks [25] [26]. Logical specifications $\theta$ may also be incorporated into the training purpose of an ANN by constructing, for example, a corresponding differentiable loss function $L(\theta)$, such that x (logically) satisfies $\theta$ whenever $L(\theta)$(x) = 0, or by incorporating constraints in a way that they will be satisfied by the model even on unseen data [27]. More generally, in case of mutual dependencies between the perception unit and the controller for realizing active perception or in case of learning-enabled controllers, sufficiently strong preconditions for these two components can be synthesized based on, for example, a combination of traditional assume-guarantee reasoning and machine learning [28] [29].

[25] Notice that the precondition $x\ \in S$ is in the language of environment inputs, whereas the postcondition is in the language of the context models.

Instead of using sets of states as properties and state transformers between these properties, one also has the possibility of building uncertainty directly into the computational model of the AI system. In these cases, the behavior of AI systems may, at least partially and when necessary, be based on probabilistic sets, where states belong to a set with a certain probability only, and probabilistic transformers between probabilistic sets. The classical notions of weakest precondition and strongest postcondition generalize to probabilistic set transformers.

Whenever there is only a small finite number of interactions between the perception and the control unit, as is expected in many real-time systems, the weakest precondition approach above is applied to the unrolled system. These kinds of pre/postcondition specifications for the perception unit are the basis for largely decoupling the development of perception from the control unit. For example, as long as the controller adapts in time such that $wp(controller')(S) \subseteq wp(controller)(S)$, where *controller'* is the updated controller, then component-wise safety analyses still compose to a system-level safety argument; otherwise the challenge is to identify corresponding minimal sets of changes for the perception unit and its analysis. The perception specification can also be used either as additional input for the training of the perception ANN or as the basis for verifying such a component; for example, for systematically deriving test cases.

These initial ideas for systematically deriving component safety, particularly for learning-enabled components from overall systems safety requirements obviously need to be further developed and stress-tested on challenging real-world AI systems.

## Component Safety Verification

Furthermore, one may compute the weakest precondition of the perception ANN. For example, computing weakest preconditions of *ReLu* networks with their rather simple node activation functions is, in principle, straightforward [30]. Now, given a safety envelope S, the safety verification problem for an AI system (perception; control) may be stated as

$$wp(perception)\big(wp(controller)(S)\big) \subseteq S$$

This fundamental safety invariant immediately reduces to the local constraint for the perception unit:

$$perception(S) \subseteq wp(controller)(S)$$

These kinds of constraints are statically analyzed based on symbolic verification techniques [31], used for test case generation, or dynamically checked by means of runtime verification (see Section 5).

Also, the perception component may now be trained with the additional knowledge that its precondition is $S$ and the postcondition is $wp(controller)(S)$. Logical constraints can also be interpreted in a more general quantitative logic for obtaining a differentiable objective function as needed for hill-climbing based training. Such quantitative interpretation may, for instance, be based on probabilistic sets and probabilistic transformers for modeling.

If we manage to train a "correct" ANN then we obtain a safety-by-design method for constructing safe AI systems. Indeed, as mentioned above, an ANN may be trained to obey some given logical safety property by constructing a corresponding differentiable loss function for the satisfiability of this formula. Still, there remains uncertainty about the safe behavior and input-output uncertainty due to the incorrectness of underlying learning algorithms. A new generation of knowledge-enhanced machine learning [32] techniques is tackling such real-world challenges for machine learning algorithms.

# 4. Uncertainty Quantification

Learning in the sense of replacing specific observations by general models is an inductive process. Such models are never provably correct but only hypothetical and therefore uncertain, and the same holds true for the predictions produced by a model.

In addition to the uncertainty inherent in inductive inference, other sources of uncertainty exist, including incorrect model assumptions and noisy or imprecise data. Correspondingly one usually distinguishes between *aleatori*c and *epistemic* sources of uncertainty [33] [34]. Whereas aleatoric[26] uncertainty refers to the variability in the outcome of an experiment which is due to inherently random effects, epistemic[27] uncertainty refers to uncertainty caused by a lack of knowledge. In other words, epistemic uncertainty refers to the ignorance of an actor, and hence to its epistemic state, and can in principle therefore be reduced with additional information. There are various approaches towards robustness based on reducing uncertainty [33]. Uncertainty reduction also plays a key role in active learning [35], and in learning algorithms such as decision tree induction [36].

Indeed, there is a multitude of sources for uncertainty in the design of safe AI systems [37]. There is, among others, uncertainty about the operational context, there is uncertainty about hazards and risks, there is uncertainty about the correctness and generalizability of learning-enabled components, there is uncertainty about safety envelopes, there is uncertainty due to noise in sensing, there is controller uncertainty due to nondeterminism and/or probabilistic control algorithms, there is uncertainty on the internal models of the controller, and, last but not least, there is also uncertainty about the actions of human operators and their possible interaction with the AI-based control system.

Rigorous approaches for safe AI need to manage the multitude and heterogeneity of sources of uncertainty. We are therefore proposing an engineering approach based on the principle of uncertainty reduction, thereby increasing predictability (up to tolerable quantities) of the AI system. Crucial steps are:

- Identification of all[28] relevant sources of uncertainty.
- Quantification and estimation of uncertainty,[29] including certainty thereof.
- Forward and inverse propagation of uncertainty along chains[30] of computation.
- Modular composition of uncertainties along architectural decomposition[31] of the AI system.

---

[26] Aka statistical, experimental, or “known unknown”
[27] Aka systematic, structural, or “unknown unknown”
[28] In a defeasible manner
[29] Uncertainty quantification is the science of quantitative characterization and reduction of uncertainties in both computational and real world applications. Among others, it tries to determine how likely certain outcomes are if some aspects of the system are not exactly known.
[30] Including recursive chains.
[31] Both horizontal and vertical.

- Design operators for mitigating overall system uncertainty below a certain level as determined by a risk and safety analysis;[32] including:
    - A combination of offline and online accumulation of relevant knowledge for managing epistemic sources of uncertainty;
    - Incremental change of uncertainty reasoning due to self-learning or even self-modification capabilities of an AI system.

Clearly, these tasks for managing the multitude of heterogeneous sources for uncertainty of AI systems are fundamental in any rigorous and transparent engineering process. We currently do not have, however, a comprehensive set of methods and tools for supporting application engineers on managing uncertainties.

## Environmental uncertainty

The operational environment of AI systems can be rather complex,[33] with considerable uncertainty even about how many and which objects and agents, both human and robotic, are in the environment, let alone about their intentions, behaviors, and strategies [38]. An AI system therefore must act without relying on a correct and complete model of the operating environment. The models at hand usually do not faithfully reflect the real-world operational context,[34] and it is simply not possible, and possibly not even desirable, to model everything. For dealing with modelling errors, AI systems may make distributional assumptions on the operational environment. It can be difficult, however, to exactly ascertain the underlying distribution.

As an alternative to explicitly modeling the operational environment it is common to specify this environment by means of a set of scenarios, which are supposed to be sampled with respect to the underlying distribution of the environment. These scenarios are analyzed and labeled with their respective interpretation of the context model in order to obtain training data for an ANN-based perception unit. It is a major challenge to select "good" scenarios. These scenarios are supposed to significantly reduce the difference between the, assumed, underlying distribution of the operating environment and the distribution of the selected training set. Collecting scenarios by driving around for, say, five hours on a stretch of highway in Alaska does not contribute as well to the approximation of real-world driving as, say, collecting driving scenarios at the Gate of India. Another concern is about evolving operating scenarios, and how to correspondingly adapt the set of specifying scenarios.

The challenge is to quantify and measure uncertainty between the operating environment and its specifying set of scenarios, identify "good" scenarios for reducing uncertainty to tolerable levels, provide sufficient conditions on the uncertainty of scenario sets for overall system safety (up to quantifiable tolerances as identified through safety risk assessment), adapt specifying scenario set to evolving operating environment.

[32] For example, less than one hazardous behavior for $10^9$ operational time
[33] Operational Design Domains may be specified following standards such as PAS 1883 (https://www.bsigroup.com/en-GB/CAV/pas-1883)
[34] Again, the old slogan applies: all models are wrong, but some might be useful.

## Behavioral Uncertainty

We restrict our considerations on learning-enabled components to the widely popular class of ANNs. Such an ANN is a deterministic function. Due to non-linear activation functions, however, there is considerable uncertainty about its input-output behavior: training instances may or may not be represented correctly by the ANN, and it is usually unclear how, and how much, the input-output behavior of an ANN generalizes from training instances. The success of one-pixel attacks serves as a reminder on the limited generalizability and resilience of some machine-learned models.

Establishing resilience [23] or invariance properties - for example, invariance with respect to certain affine or homeomorphic transformations - of an ANN is an important means for reducing uncertainty on the input-output behavior. Some uncertainty about outcomes, however, remain. A systematic framework for analyzing different sources of uncertainty for ANNs is described in [39].

*Measuring behavioral uncertainty.* Entropy may be used for quantifying uncertainty of a neural network. Indeed, under some mild assumptions on uncertainty, entropy is the only possible definition of uncertainty [40], at least in its aleatoric interpretation. There are a multitude of indicators of behavioral uncertainty. [41], for example, proposes to use the distance between neuron activations observed during training and the activation pattern for the current input as an estimation for the uncertainty of the input-output behavior.

*Training-based estimation of behavioral uncertainty.* Ensembles of neural networks, for example, estimate predictive uncertainty by training a certain number of NNs from different initializations and sometimes on differing versions of the dataset. The variance of the ensemble’s predictions is interpreted as its epistemic uncertainty. Instances of ensemble learning techniques such as Bayesian neural networks (BNN) [42] measure both epistemic uncertainty $P(\theta|D)$, on model parameters $\theta$ and the aleatoric uncertainty $P(Y|X, \theta)$. In fact, the predicted uncertainty of Bayesian neural networks (BNN) is often more consistent with observed errors, compared to classical neural networks. Out-of-training distribution points of a BNN leads to high epistemic uncertainty. The uncertainty $P(\theta|D)$ can be reduced with more data. BNNs are also an interesting approach for active learning, as one can interpret the model predictions and see if, for a given input, different probable parametrizations lead to different predictions. In the latter case, labelling of such an input will effectively reduce the epistemic uncertainty.

## Uncertainty Propagation

What we really should care about is not freedom from *faults* but absence of *failure* [43]. Particularly, if a perception unit fails to meet its safety specification then we call this unit *faulty,* and if the overall cognitive system loop fails to act safely then there is a system *failure*. Using corresponding random variables $Faulty$ and $Failure$ we are interested in the probability that the system is safe, that is $P(not\ Failure)$; using Bayes' rule we obtain:

$$P(Failure \mid Faulty) * P(Faulty) = P(Faulty \mid Failure) * P(Failure)$$

Provably worst-case distributions [44] are used for estimating the posterior probability $P(Failure \mid Faulty)$ of faulty behaviors leading to safety violations.[35] The probability $P(Faulty)$ that the perception unit is faulty is approximated, for instance, using training-based estimation of behavioral uncertainty (as described above) or, alternatively, from an assurance-based estimation of uncertainty (as described below). Now, assuming that all but the perception unit are *possibly perfect* and that the faulty perception unit is the only possible cause of failure, then $P(Faulty|Failure) = 1$. Consequently, we can estimate $P(not\ Failure) = 1 - P(Failure)$ by means of Bayesian inference.

This short exposition of propagation of component faults to system safety failures is intended to demonstrate a possible style of Bayesian inference for establishing safety results. The underlying methodology however should also be applicable for more general mutually recursive system architectures.

## Assurance-based uncertainty estimation

The goal of rigorous design is to gain sufficient confidence that failures, in our case safety violations, are very rare, up to tolerable quantities. Sufficient confidence cannot, however, be built up by looking at failures only.

Instead, assurance builds up a convincing case that failures are rare. One widely quoted definition for the corresponding notion of a safety case comes from [45]: "A safety case is a structured argument, supported by a body of evidence that provides a compelling, comprehensible and valid case that a system is safe for a given application in a given operating environment." An assurance case is simply the generalization of a safety case to properties other than safety.

An assurance case therefore is a comprehensive, defensible, and valid justification of the safety of a system for a given application in a defined operating context. It is based on a structured argument of safety considerations, across the system lifecycle, which can assist in convincing the various stakeholders that the system is acceptably safe.

[35] *Failure* and *Faulty* are random variables, and the conditional probability *P(Failure | Faulty)* measures the uncertainty that the system is unsafe (*Failure*) given that the perception unit violates its specification *(Faulty).*

The purpose is, broadly, to demonstrate that the safety-related risks associated with specific system concerns[36] have been identified, are well-understood, and have been appropriately mitigated, and that there are mechanisms in place to monitor the effectiveness of safety-related mitigations. In this sense, an assurance case is a structured argument for linking safety-related claims through a chain of arguments to a body of the appropriate evidence. One of the main benefits for structured arguments in assurance cases is to explicitly capture the causal dependencies between claims and the substantiating evidence.

Altogether, assurance cases are the basis for judging that a technical system is acceptable for widespread use. Assurance cases also determine the level of scrutiny needed for developing and operating systems which are acceptably safe. More specifically, assurance cases determine constraints on the design, implementation, verification, and training strategies, and they demonstrate the contributions of corresponding artifacts and activities to the overall system safety.

One may be confident in such an assurance based on "the quality of state of being certain that the assurance case is appropriately and effectively structured, and correct" [46]. A necessary aspect of gaining confidence in the assurance case is dealing with uncertainty, which, as we have seen above, may have several sources. Uncertainty, often impossible to eliminate, nevertheless undermines confidence and must therefore be sufficiently bounded.

Recent extensions of assurance cases for reasoning about confidence and uncertainty [47] are a good starting point for estimating and managing aleatoric and epistemic uncertainties for safe AI systems. In particular, probability theory has been proposed for quantifying confidence and uncertainty ( [48], epistemic uncertainty is quantified through the Dempster-Shafer theory of beliefs or Bayesian analysis [49] , the use of Bayesian Belief Networks [50] [51] [52], Josang's opinion triangle [47], evidential reasoning [53], and weighted averages [54].

There is, however, a slight problem with quantifying confidence in assurance case arguments, as proposed methods on Bayesian Belief Networks, Dempster-Shafer, and similar forms of evidential reasoning can deliver implausible results [55, 46]. Without strong evidence that the quantified confidence assessments are indeed trustworthy, there is no plausible justification for relying on any of these techniques in safety engineering. Alternatively, one may also look towards a value for the *probability of perfection* - based on extreme scrutiny of development, artifacts, and code - which is then related to confidence [56] [57].

Qualitative approaches towards uncertainty, on the other hand, focus on the reasoning and rationale behind any confidence by building up an explicit confidence argument. For example, eliminative induction is increasing confidence in assurance cases by removing sources of doubt and using Baconian[37] probability to represent confidence [58]. Eliminative induction first identifies potential sources of doubt, so-called *defeaters*, and then work towards removing each source of doubt or proving that it is not relevant. The search for defeaters, and their own possible defeat, should be systematized and documented as essential parts of the case [59]. One systematic approach is through construction and dialectical consideration of counterclaims and countercases. Counterclaims are natural in *confirmation measure*s as studied in Bayesian confirmation theory, and countercases are assurance cases for negated claims.

[36] Including safety and security, but also applies to all the other attributes of trustworthiness.

Assurance cases have successfully been applied to many safety-critical systems, and they have also proven to be flexible enough to be adopted to systems with learning-enabled components. An overall assurance framework for AI systems with an emphasis on quantitative aspects, e.g., breaking down system-level safety targets to component-level requirements and supporting claims stated in reliability metrics has recently been outlined [60]. Requirements on assurance cases for autonomously acting vehicles with learning-enabled components are addressed, for example, by UL 4600.[38]

A mixture or requirements and data-centric metrics together with corresponding verification techniques, both static and dynamic [61], is needed to establish safety of AI systems with machine learning components A successful element in a successful deployment of safety assurance for AI systems is a library of pre-validated argument steps [62] [63, 64] together with adequate operators for instantiating and composing specific system-specific assurance cases from these pre-validated structured arguments. We also hypothesize that, due to the multitude of sources of uncertainty, assurance arguments for increasingly autonomous AI systems, need to (1) stress rigor in the assessment of the evidence and reasoning employed, and (2) systematize and automate the search for defeaters, the construction of cases and counter cases, and the management and representation of dialectical examination. Increased rigor and automation in building and maintaining assurance cases should enable productive interaction with tools for logical and probabilistic reasoning and formal argumentation. Using frameworks such as STPA [65] to better capture and examine a component's control actions in relation to the larger system level safety contexts may be beneficial. It is of particular interest how the influence of learning-enabled components is captured and reasoned within the AI control structure. And, finally, rigorous assurance cases open new possibilities of online self-adaptation of safety arguments for determining safe behavior when operating in uncertain contexts, since they can be adapted, quickly and efficiently, to ever-changing safety considerations of AI systems.

---

[37] https://ntrs.nasa.gov/api/citations/20160013333/downloads/20160013333.pdf

[38] https://ul.org/UL4600

# 5. Analysis

A key issue for AI systems is rigorous safety analysis, which is based on a mixture of well-known verification and validation techniques with safety verification of learning-enabled components. Here we focus on novel aspects on analyzing AI systems with ANN-based perception units only.

But what do we actually need to verify about ANN components in order to support AI system safety? Our starting point here are the component requirements as obtained by breaking down application-specific systems safety requirements to verification and validation (V&V) requirements on the individual components of AI systems.

Due to mounting concerns of using ANNs for safety-related applications there has been a proliferation of new techniques with the intent of increasing their trustworthiness [66] [67] [68] [64]. It is well beyond the scope of these notes to survey this ever-growing number of methods and technologies for increasing the trustworthiness of ANN. Indeed, there is no lack of individual methods, but the safety relevance of many ANN analysis techniques (such as adversarial analysis) is questionable, particularly when the impact of the overall system within which the ANN is used is unclear [69].

What is needed is a systematic evaluation of individual analysis techniques. A central challenge is to adequately measure and quantify how well and under which circumstances they improve confidence in the safe system behavior. A first step in this direction is provided in [70], as they develop a safety pattern for choosing and composing analysis techniques based on how they contribute to the identification and mitigation of systematic faults known to affect system safety. More generally, given an ANN and some desired properties, we therefore define the goal of ANN analysis to improve confidence, or, dually, reduce uncertainty, whether the desired properties hold, up to tolerable quantities, on the ANN.

## Testing

The goal of testing ANNs is to generate a set of test cases, that can demonstrate confidence in an ANN's performance, when passed, such that they can support an assurance case. Usually, the generation of test cases is guided by coverage metrics, both structural and non-structural [71].

Traditional structural coverage criteria from software testing can usually not be applied directly to ANN. For example, neuron coverage is trivially fulfilled in ANN by a single test case. Moreover, MC/DC, when applied to ANNs, may lead to an exponential (in the number of neurons) number of branches to be investigated, and are therefore not practical as typical ANNs are comprised of millions of neurons. As usual in testing, the balance between the ability to find bugs and the computational cost of test case generation is essential for the effectiveness of a test method [72].

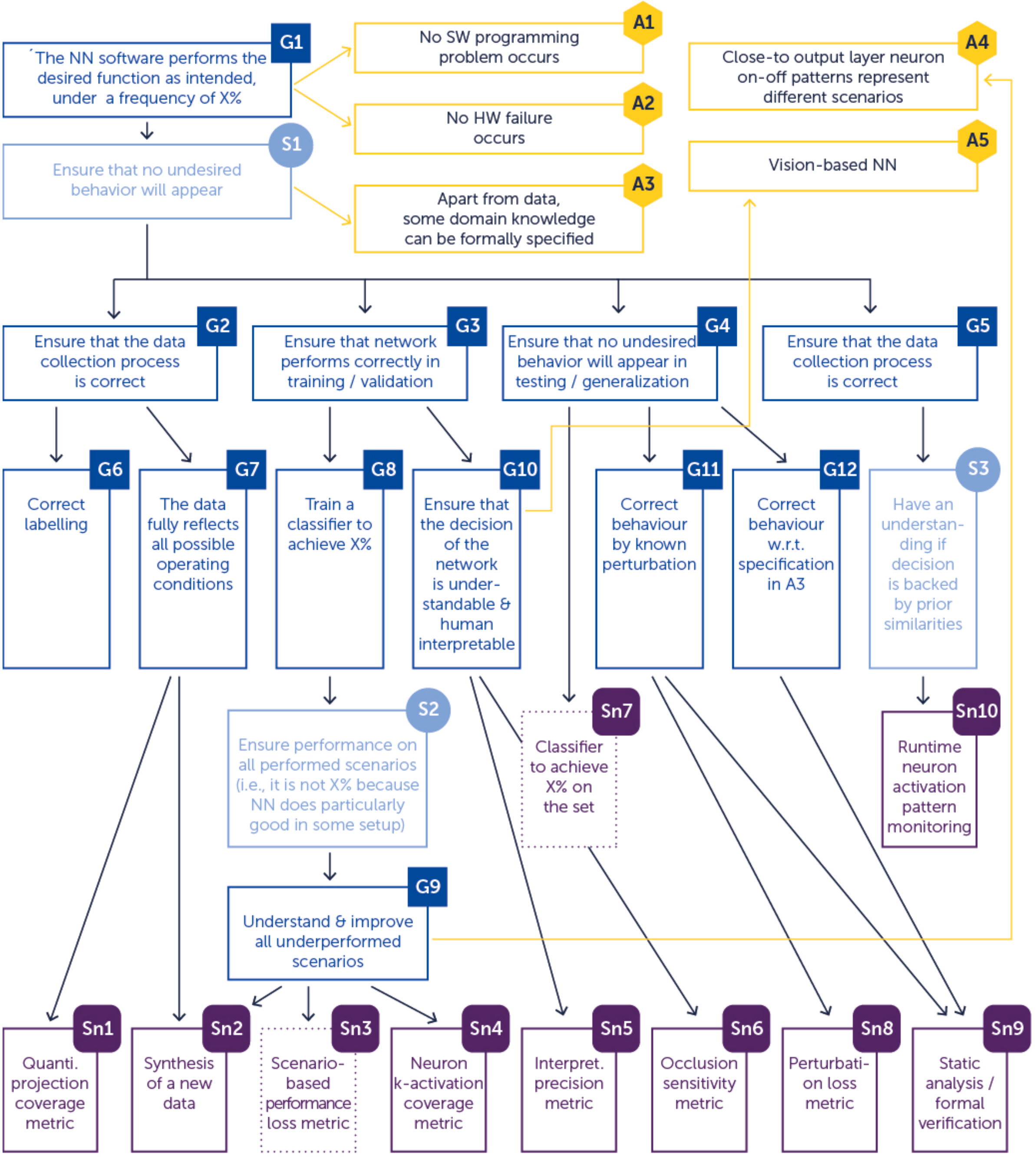

*Figure 4.* NNDK-based assurance case.

The generation of falsifying/adversarial test cases is generally using search heuristics based on gradient descent or evolutionary algorithms [73, 74] [75] [76]. These approaches may be able to find falsifying examples efficiently, but they usually do not provide an explicit level of confidence about the nonexistence of adversarial examples when the algorithm fails to find one.

[20] develop ANN-specific non-structural test coverage criterion for the robustness, interpretability, completeness, and correctness of an ANN. A scenario coverage metric, for example, partitions the possible input space according to N attributes (e.g. snow, rainy, …), and proposes, based on existing work on combinatorial testing, efficient k-projection (for k = 0,…,N-1) coverage metrics as approximations of the exponential number of input partitions. In principle, a “complete” (with respect to the available input data) set of attributes may be obtained through unsupervised learning or clustering methods. These coverage metrics are implemented in the NNDK testing toolkit for ANNs [77].

In [78] coverage is enforced to finite partitions of the input space, relying on predefined sets of application-specific scenario attributes. In a similar vein, the “boxing clever” technique focuses on the distribution of training data and divides the input domain into a series of representative boxes.

A number of traditional test case generation techniques such as fuzzing [79, 80, 75] [81], symbolic execution [82], concolic testing [83], mutation testing [84], mutation testing, and metamorphic testing [85] have been extended to support the verification of ANNs. Despite their effectiveness in discovering various defects of ANNs together with their data-centric requirement specifications, however, it is not exactly clear how testing-based approaches can be efficiently integrated into the construction of convincing safety argumentations for AI systems. A possible step in this direction, however, is the NNDK-based safety case in Figure 4, which makes the contribution and the rationale behind individual test metrics in establishing safety goals more explicit.

Altogether, testing methods seem to be effective at discovering defects of ANNs. It is unclear, however, how to measure the effectiveness of test coverage metrics in building up sufficient confidence – or dually, raising doubts – in a convincing assurance case. Also, most testing-based approaches assume a fixed ANN. However, ANNS are learning-enabled and trained continuously on new data/scenarios. The challenge is to come up with methodologies for efficiently - depending on the application context also in real-time - retesting safety requirements for continuously evolving ANNs. Such a retesting methodology could be based on adapting corresponding assurance cases.

Instead of validating individual learning-enabled components, the idea of scenario-based testing is to (1) automatically or manually identify a reasonably small set of relevant dynamic situations, or scenario types; (2) check if the set of scenario types is complete; and then (3) derive system-specific tests for each scenario type. This immediately raises the need for a test ending criterion based on the following question: did we test all scenario types? And did we sufficiently test each type with specific instances? The general approach to scenario-based testing is outlined in Figure 5. It is based on automated clustering of real driving data and completeness checks for the clusters thus obtained [86].

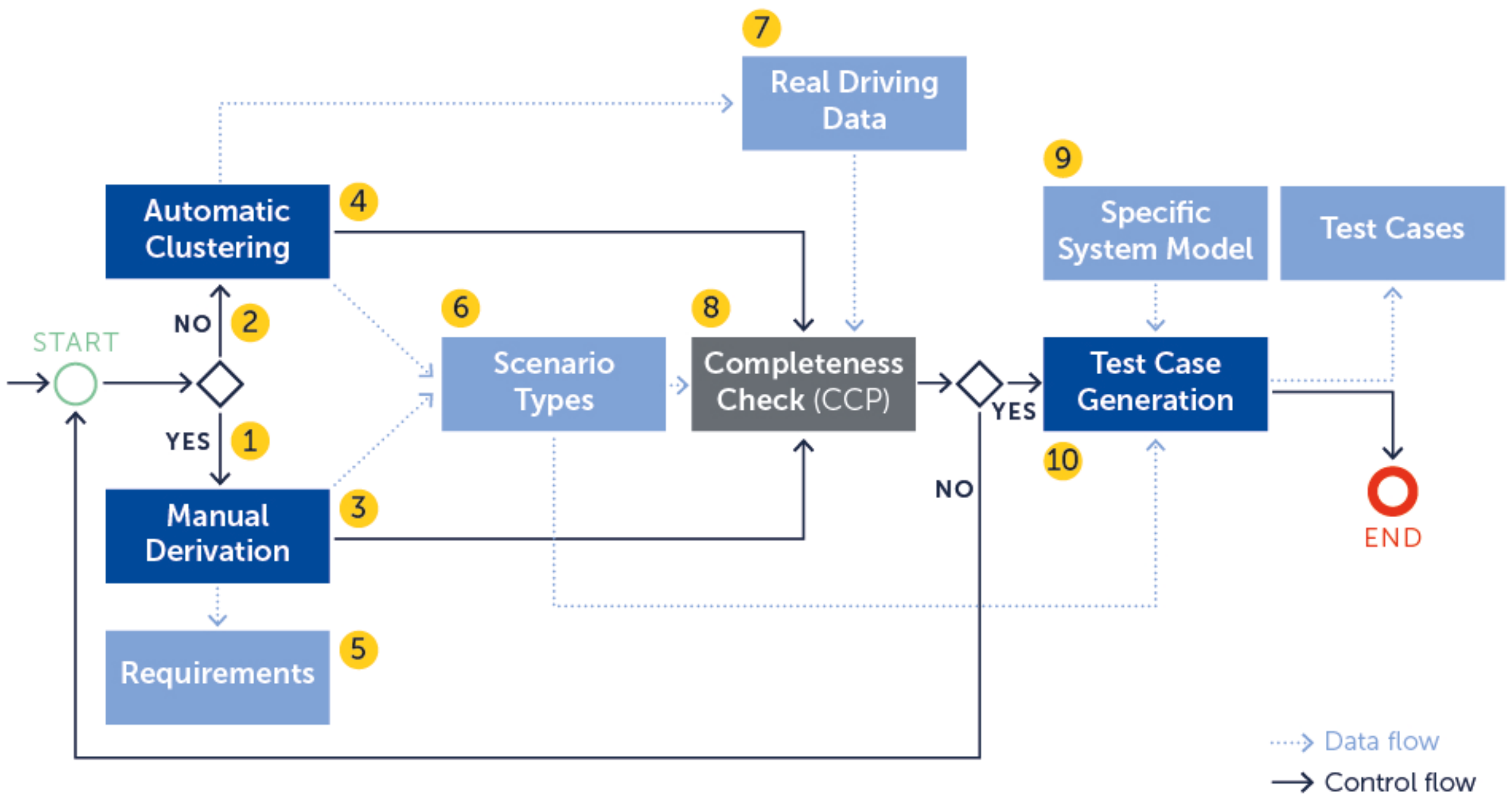

***Figure 5.*** Scenario-Based Testing.[39]

## Symbolic Verification

Safety verification problems for ANNs can be reduced to constraint solving problems such as satisfiability in propositional logic [87] [88], satisfiability modulo theories [89] [90] [91] [92], and mixed-integer linear programming [23]. These approaches typically do not scale up to the size of real-world ANNs with millions of neurons. Approximation techniques are applied to improve efficiency, but usually at the expense of precision. Recent approaches based on global optimization have the potential of dealing with larger networks [93]. Compositional verification techniques for scaling up ANN safety verification are largely missing. For assume-guarantee style reasoning applied to the verification of an ANN-based automotive safety controller, however, see [94].

Since symbolic safety verification technologies work on a model of the ANN they might certain defects due to implementation issues (for example, rational numbers vs. IEEE floating point implementations). It is also unclear how to efficiently apply these techniques to continuously changing ANNs.

[39] Adapted from: https://doi.org/10.1109/ITSC.2019.8917326

## Runtime verification

In runtime verification a monitor observes the concrete execution of the system in question and checks for violations of stipulated properties. When the monitor detects a violation of a property, it notifies a command module which then isolates the cause of the violation and attempts to recover from the violation. In this way, runtime verification is a central element of FDIR-based[40] fault-tolerant systems. For the multitude of sources for uncertainty in AI systems, stringent real-time requirements, and ever-changing learning-enabled components, runtime verification is an essential element for safety verification of AI systems.

System requirements of the form "the system must perform action *a* within *n* seconds of event *e*" are common in runtime monitoring of autonomous systems [95]. These kinds of properties are expressible in suitable sub-logics of metric temporal logic such as GXW [96] [97, 52] and timed extensions thereof [98]. These kinds of specifications are compiled into (timed) synchronous dataflows as the basis for efficient runtime monitors. A dynamic programming and rewriting-based algorithm for monitoring MTL formulas is described in [99]. Moreover, architectural design principles for monitoring distributed systems are needed to ensure that monitoring does not perturb the system (at least, not too much) [74]. In particular, the tutorial [100] discusses challenges on instrumenting real-time systems so that timing constraints of the system are respected. A recent tutorial describes state-of-the-practice technology for generating runtime monitors that capture the safe operational environment of systems with AI/ML components [101].

Altogether, runtime verification is an essential and attractive technique of any verification strategy for safe AI. Unlike static verification techniques such as testing or symbolic verification, there is no need for adaptation to learning-based components such as ANNs. In this way, runtime monitoring is an enabling verification technology for continuous assurance, based on the MAPE-K[41] loop from autonomic computing. The main challenge in deploying runtime monitoring, as is the case for any other cyber-physical system, is to embed monitors in an efficient (for example, energy-efficient) way, without perturbing the behavior of the AI system too much.

Runtime monitoring may also be used for measuring uncertainties in input-output behavior of ANNs. For example, if an input is out-of-distribution of the training set, then one may conclude that the "correctness" of the corresponding ANN output may be doubtful. Such information about the uncertainty of a perception result may be useful input for planning in the deliberation stage. Uncertainty information about the perception unit is also used in Simplex architectures for switching to a safe(r) perception channel whenever the ANN output is doubtful. Clearly, the distance (in some given metric) of the input to the set of training input may serve as a measure of uncertainty of the input-output behavior of the ANN. Notice, however, that such a measure returns uncertainty zero even for “incorrect” behavior of the ANN on training inputs. Alternatively, [102] propose to monitor the neuronal activation pattern of some input, and to compare it with neuronal activation patterns as learned during the ANN training phase. Notice that this measure of certainty on the input-output behavior of an ANN is part of the assurance case for an ANN in Figure 4. In addition, applicable background knowledge and physical laws may also be used in monitoring the plausibility of the input-output behavior of an ANN.

[40] **F**ault **D**etection, **I**solation, and **R**ecovery.

[41] **M**easure, **A**nalyze, **P**lan, **E**xecute; the **K** stands for Knowledge.

In summary, due to the multitude of sources of uncertainty, the complexity of AI-based systems and the environments in which they operate, even if all the challenges for specification and verification are solved, it is likely that one will not be able to prove unconditional safe and correct operation. There will always be situations in which we do not have a provable guarantee of correctness. Therefore, techniques for achieving fault tolerance and error resilience at run time must play a crucial role. There is, however, not yet a systematic understanding of what can be achieved at design time, how the design process can contribute to safe and correct operation of the AI system at run time, and how the design-time and run-time techniques can interoperate effectively.

# 6. Safety-by-Design

Validation and verification activities are usually complemented with *safe-by-construction* design steps. We briefly describe some of the main challenges and initial approaches towards safe-by-design; namely property-driven synthesis of learning-enabled components, compositional construction of AI subsystems and systems, and safety architectures for AI systems. The goal in this respect is a fundamental set of building blocks together with composition and incremental change operators for safe-by-construction design and for continual assurance of large classes of AI systems.

## Property-driven synthesis

Instead of using *a posteriori* verification of desirable properties of ANNs by means of static or dynamic verification technologies as outlined above, it is a natural question to ask ourselves: can we design, from scratch, a machine learning component that provably satisfies (possible in a robust interpretation) given formal specifications? For example, given pre- and postconditions of an ANN, as obtained from breaking down system safety envelopes down to individual learning-enabled components, is it possible to train an ANN that satisfies the given safety specification. Given a property expressed in logic, for example, one constructs a corresponding differentiable loss function for property-driven training of the ANN. In this way, property-driven synthesis needs to, among other things, design an appropriate training set, set up the initial structure of the ANN, choose and adjust appropriate hyper-parameters for training. The selection of training sets and training then is guided by reducing an adequate measure of the uncertainty that the ANN indeed satisfies the given specification.

Progress is needed along all these fronts. Techniques of neuro-symbolic computation [103] [104] may be a good starting point, as they also try to integrate high-level reasoning with low-level perception in such a way that neuro-symbolic methods have the pure neural, logical, and probabilistic methods as special cases. A short history and perspectives for knowledge-augmented machine learning are described in [32].

## Compositional System Design

The triad of perception, deliberation, and execution as depicted in Figure 1 is only the simplest possible architecture of an AI system. Often, deliberation and execution units are complex and mutually dependent for realizing a fine-grained control, perception may also be dependent on deliberation, say, in AI systems with active perception. Moreover, each stage of the AI systems triad is usually decomposed into any number of functional units including monitors and safe channels. For example, deliberation may include functionalities for modeling AI capabilities such as interpretation and prediction, model building, derivation of knowledge, goal management, or planning, and perception is decomposed into a pipeline of tasks for, say internal and external state estimation, sensor fusion, object recognition, object classification. Such a real-world architecture for realizing an autonomous driving function can be found, for example, in [105].[42]

Traditional Simplex architectures [106] are used for addressing both performance and safety requirements of a number of automated and autonomous systems [107] [108] [109] [110] by leveraging run-time assurance, where the results of design-time verification are used to build a system that monitors itself and its environment at run time. More precisely, a Simplex architecture comprises (1) a performant controller under nominal operating conditions, which is design to achieve high-performance, but it is not provably safe (2) a safe controller that can be pre-certified to be safe, and (3) a decision module which is pre-certified (or safe-by-design) to monitor the state of the controlled system and its operational environment to check whether desired system safety specifications can be violated. If so, the decision module switches control from the nominal to the safe monitor. Provably safe composition of Simplex architectures is developed in the context of *Soter* [111], which also allows for switching to nominal control as to keep performance penalties to a minimum while retaining strong safety guarantees.

While compositional design operators have been developed for digital circuits and embedded systems, we do not yet have such comprehensive theories for AI systems. For example, if two ANNs are used for perception on two different types of sensors, say LiDAR and camera, and individually satisfy their specifications under certain assumptions, under what conditions can they be used together for decreasing perception uncertainty? More generally, how can we compositionally design safe and predictable perception pipelines? How can one design planning and deliberation components for overcoming the inherent limitations of their ANN-based perception component? How can one design execution components for minimizing surprises in uncertain environments? And how can these components interact in a safe and quasi-predictable[43] manner?

[42] This publication advocates the use of state-of-the-practice dependability and safety engineering methodologies as prescribed in current industrial safety standards,[42] for safing SAE L3 and L4 automated driving capabilities.

[43] That is, predictable up to acceptable levels.

| | |
|---|---|
| **Specification Challenge** | • Provide means for constructing (and maintaining) safety envelopes, either deductively from safety analysis or inductively from safe nominal behavior<br>• Provide means for minimizing uncertainties related to safety envelopes with a given level of effort<br>• Provide means for deriving safety requirements for learning-enabled components, which are sufficient for establishing AI system safety<br>• Provide means for reducing specification uncertainty by means of deriving data requirements for learning-enabled components |
| **Uncertainty Challenge** | • Identify all relevant sources of uncertainty for an AI system<br>• Provide adequate means for measuring uncertainty<br>• Calculate forward propagation of uncertainty, where the various sources of uncertainty are propagated through the model to predict overall uncertainty in the system response<br>• Identify and solve relevant inverse[44] uncertainty quantification problems for safe AI<br>• Predict (up to tolerable quantities) unsafe behavior of AI systems operating in uncertain environments |
| **Assurance Challenge** | • Provide adequate measures of uncertainty for assuring AI system safety<br>• Construct and maintain evidence-based arguments for supporting the certainty and for rebuting the uncertainty of safety claims<br>• Identify useful safety case patterns[45] for safe AI systems and identify corresponding operators for instantiating these patterns and for composing them |
| **Design Challenge** | • Develop safety case patterns for different architectural designs of AI systems[46]<br>• Compositionally construct safe and quasi-predictable AI systems together with their safety cases |
| **Analysis Challenge** | • Provide adequate means for measuring and for reducing uncertainty on the input-output behavior of learning-enabled components<br>• Define and measure the respective contribution of static and dynamic analysis techniques for learning-enabled systems, towards reducing safety-related uncertainty to tolerable levels |
| **Maintenance Challenge** | • Identify incremental change operators for maintaining uncertainty and safety assurance of self-learning AI systems<br>• Safely adapt and optimize the situational behavior of an AI system (together with its safety cases based on the principle of minimizing uncertainty |

***Table 1.*** Safe AI Engineering Challenges.

[44] That is, calculating from a set of observations the causal factors that produced them
[45] Cmp. AMLAS
[46] *In analogy to, say, Mils separation kernel protection profile.*

# 7. Conclusions

We have been arguing that traditional safety engineering is not suitable for developing and operating AI systems for safety-related applications. Based on this insight we outlined a safety engineering methodology for AI, which is centered around managing and assuring uncertainty to acceptable levels [112, 76], as the basis for predicable (up to acceptable tolerances) and safe AI systems.

The proposed rigorous design methodology for safe AI is based on the central notion of a safety case for managing uncertainties. Our proposals are compatible with the emergent standard UL4600[47] on required properties for safety cases. In some sense, the depicted design methodology may also be viewed as an uncertainty-based amalgam of the paradigms of data- with model-driven design.

The main contribution lies in the identification of core challenges and possible research directions on the specification, design, analysis, assurance, and maintenance of safe AI (for a summary, see *Table 1*). This list, however, is incomplete, as we have omitted, for instance, all-important challenges due to interactive control between human operators and the machine-based control.

The identified challenges for safe AI as listed in *Table 1* do not seem to be insurmountable hard to overcome. The overarching challenge rather lies in the integration of individual methods into a coherent and comprehensive engineering framework for systematically managing and reducing uncertainty to tolerable quantities, and to demonstrate its relative merits on real-world AI systems.

We have been working towards AI safety engineering, among others, with Fasten ( [113, 20] for checkable safety cases, evidential transactions in Evidentia/CyberGSN for continual assurance and compliance [114], and the neural network dependability kit [77] for analyzing artificial neural networks, and risk-based safety envelopes for autonomous vehicles under perception uncertainty [115]. We are also currently working on concrete safe AI use cases for integrating these individual engineering nuggets, and for elaborating a generally useful approach for safe AI engineering. We hypothesize that uncertainty quantification also plays an increasingly prominent role in analyzing and certifying complex software systems, since traditional notions of system-level correctness are becoming less applicable for heterogeneous and ever-evolving software landscapes.

There are related ideas on *uncertainty quantification* in engineering [80] for certifying that, with high probability, a real-valued response function of a given physical system does not exceed a given safety threshold. Uncertainty quantification also plays a pivotal role in minimizing uncertainties for ANNs [116]. We expect these kinds of techniques to provide a mathematical underpinning of a design calculus for safe AI.

[47] https://edge-case-research.com/ul4600/

The ultimate goal in this respect is a rigorous engineering framework based on pre-certified parameterized components, corresponding assurance arguments, and system composition operators (for example, for watchdogs, monitors, and redundant channels) from which complete systems and corresponding assurance cases are constructed in a property-guided, traceable, and optimized manner.

In addition, onboard management of uncertainty is used for the design of safe exploration strategies of unknown territory based on the principle of managing uncertainty and for minimizing surprises. This kind of safe exploration of an AI system might even be complemented with online risk and safety assessment together with corresponding online updates of safety cases and uncertainty quantifications thereof.